\documentclass[11pt,a4paper]{article}
\usepackage[hyperref]{acl2020}
\usepackage{times}
\usepackage{latexsym}
\usepackage{graphicx}
\usepackage{url}
\usepackage{amsmath}
\usepackage{float}
\usepackage{xcolor}
\usepackage{subcaption}
\usepackage{multirow}
\usepackage{amsfonts}
\usepackage{booktabs}
\usepackage{bbm}
\usepackage{pifont}
\usepackage{makecell,rotating}
    \setcellgapes{5pt}

\hypersetup{colorlinks,urlcolor={blue}}

\aclfinalcopy % Uncomment this line for the final submission
\def\aclpaperid{136} %  Enter the acl Paper ID here

\newcommand{\minus}{\scalebox{0.55}[1.0]{$-$}}

\title{
SCDE: Sentence Cloze Dataset with High Quality Distractors From Examinations
}
\newenvironment{tight_enumerate}{
\begin{enumerate}
  \setlength{\itemsep}{0pt}
  \setlength{\parskip}{0pt}
}{\end{enumerate}}
\newenvironment{tight_itemize}{
\begin{itemize}
  \setlength{\itemsep}{0pt}
  \setlength{\parskip}{0pt}
}{\end{itemize}}
\author{Xiang Kong\bf{\thanks{\quad Equal Contribution}}, Varun Gangal\footnotemark[1], Eduard Hovy \\
        Language Technologies Institute \\ Carnegie Mellon University \\ {\tt \{xiangk,vgangal,hovy\}@cs.cmu.edu } }
\date{}
\newcommand{\dsname}{\textsc{SCDE}}
\begin{document}
\maketitle
\begin{abstract}
We introduce \dsname, a dataset to evaluate the performance of computational models through sentence prediction. \dsname~is a human-created \emph{sentence} cloze dataset, collected from public school English examinations. Our task requires a model to fill up multiple blanks in a passage from a shared candidate set with distractors designed by English teachers.
Experimental results demonstrate that this task requires the use of \emph{non-local}, \emph{discourse-level} context beyond the immediate sentence neighborhood. The blanks require \emph{joint} solving and significantly impair each other's context. Furthermore, through ablations, we show that the distractors are of high quality and make the task more challenging. Our experiments show that there is a significant performance gap between advanced models (72\%) and humans (87\%), encouraging future models to bridge this gap.\footnote{Data: \url{vgtomahawk.github.io/sced.html}} \footnote{Code: \url{https://github.com/shawnkx/SCDE}}

\end{abstract}
%Just made it Equal Contribution. Earlier sentence is too long and people may skip it.

\section{Introduction}
\label{sec:intro}
Cloze questions were first proposed by \newcite{taylor1953cloze} as a readability test, motivated by Gestalt psychology. They become an efficient way of testing reading for public exams, overtaking the dominant paradigm of subjective questions ~\cite{fotos1991cloze,jonz1991cloze}. 
Cloze datasets~\cite{zweig2011microsoft,hermann2015teaching,hill2015goldilocks,paperno2016lambada,onishi2016did,xie2017large} became prevalent as question-answering (QA) benchmarks since they are convenient either to be generated automatically or by annotators.  These datasets could be split into two clear types:
\begin{enumerate}
    \item Where the context is a complete text, and there is an explicit question posed which is a statement with a cloze gap. The answer is either generated freely or is a span from the context, e.g. Children’s  Books Test (CBT)~\cite{hill2015goldilocks}.
    \item Where the context itself comes with cloze gaps. There is no explicit question. The answer is generated freely or chosen from a set of candidates, e.g. CLOTH~\cite{xie2017large}.
\end{enumerate}

\begin{table*}[!ht]
\small
	\centering
	\fbox{\begin{minipage}[t]{445pt}
{\footnotesize
{\bf Passage:}

A student's life is never easy. And it is even more difficult if you will have to complete your study in a foreign land. ${\bf \rule{1cm}{0.15mm}}_{1}$ The following are some basic things you need to do before even seizing that passport and boarding on the plane.  Knowing the country. You shouldn't bother researching the country's hottest tourist spots or historical places. You won't go there as a tourist, but as a student. ${\bf \rule{1cm}{0.15mm}}_{2}$ In addition, read about their laws. You surely don't want to face legal problems, especially if you're  away from home. ${\bf \rule{1cm}{0.15mm}}_{3}$ Don't expect that you can graduate abroad without knowing even the basics of the language. Before  leaving your home country, take online lessons to at least master some of their words and sentences. This will be useful in living and studying there. Doing this will also prepare you in communicating with those who can't speak English. Preparing for other needs. Check the conversion of your money to their local currency. ${\bf \rule{1cm}{0.15mm}}_{4}$. The Internet of your intended school will be very helpful in findings an apartment and helping you understand local currency. Remember, you're not only carrying your own reputation but your country's reputation as well. If you act foolishly, people there might think that all of your countrymen are foolish as well. ${\bf \rule{1cm}{0.15mm}}_{5}$

{\bf Candidates:}

A. Studying their language.

B. That would surely be a very bad start for your study abroad program. 

C. Going with their trends will keep it from being too obvious that you're a foreigner.

D. Set up your bank account so you can use it there, get an insurance, and find an apartment.

E. It'll be helpful to read the most important points in their history and to read up on their culture.

F. A lot of preparations are needed so you can be sure to go back home with a diploma and a bright future waiting for you.

G. Packing your clothes.

{\bf Answers with Reasoning Type:}

1$\rightarrow$F (\textit{Summary}) , 2$\rightarrow$E (\textit{Inference}) , 3$\rightarrow$A (\textit{Paraphrase}) , 4$\rightarrow$D (\textit{WordMatch}), 5$\rightarrow$B (\textit{Inference}) (C and G are distractors)

{\bf Discussion:}

Blank 3 is the easiest to solve, since ``Studying their language'' is a near-paraphrase of ``Knowing even the basics of the language''. Blank 2 needs to be reasoned out by \textit{Inference} - specifically $E$ can be inferred from the previous sentence. Note however that $C$ is also a possible inference from the previous sentence - it is only after reading the entire context, which seems to be about learning various aspects of a country, that $E$ seems to fit better. Blank 1 needs \textit{Summary} $\rightarrow$ it requires understanding several later sentences and abstracting out that they all refer to \emph{lots of preparations}. Finally, Blank 5 can be mapped to B by inferring that \textit{people thinking all your countrymen are foolish} is \emph{bad}, while Blank 4 is a easy \textit{WordMatch} on \textit{apartment} to D. The other distractor G, although topically related to preparation for going abroad, does not directly fit into any of the blank contexts
}
\end{minipage}}
\caption{A Representative Example from \dsname.}
\label{tab:mainExample}
\vspace{-2ex}
\end{table*}

Herein, we focus on the 2nd category. A common property of these datasets is that they have gaps at the level of words, entities or short syntactic spans. The entity and span-based clozes may sometimes be multi-token, but they do not extend beyond a few tokens. Nevertheless, none of these datasets have cloze gaps at the level of \emph{full sentences}. Since many syntactic and semantic cues are present in the same sentence, this makes the gap easier to fill compared to the sentence level cloze case where models would have to rely on ``discourse" cues beyond the same sentence.

Besides lack of intra-sentence cues, sentence-level cloze may require comparing candidates of very different lengths. For instance, the example in Table \ref{tab:mainExample} has a standard deviation of $7.6$ with candidate lengths between $3$ to $25$. A model that only represents words well may not get comparable probabilities at sentence level for very different sentence lengths. Therefore, robust sentence representation models are also required to solve this question. 
In this paper, we present \dsname, a dataset of sentence-level cloze questions sourced from public school examinations. Each dataset example consists of a passage with multiple sentence-level blanks and a shared set of candidates. Besides the right answer to each cloze in the passage, the candidate set also contains ones which don't answer any cloze, a.k.a., \emph{distractors}. Both cloze positions and distractors are authored by teachers who design the public school examinations carefully. \S \ref{sec:dataCollection} explains our data collection. A representative example from \dsname~is shown in Table \ref{tab:mainExample}. 

Another salient aspect of our dataset is that more than 40\% of blanks belong to the reasoning category \emph{``Inference"} (more on this in \S \ref{subsec:reasoningType} and Table \ref{tab:categoriesTwo}) which require models to compare  plausibility of competing hypotheses given a premise (whether the previous or last sentence(s), or even a combination of information from the two). Filling these blanks requires the model to reason by using commonsense knowledge, factual knowledge, time gaps, etc. Some of these can be thought of as simple entailment, but more generally, many of these can be seen as requiring abductive reasoning, which is of recent interest \cite{bhagavatula2019abductive,sap2019atomic,sap2019socialiqa} to the NLP community. In summary, our contributions are as follows
\vspace{-\topsep}
\begin{tight_enumerate}
  \setlength{\parskip}{0pt}
  \setlength{\itemsep}{0pt plus 1pt}
   \item We introduce the task of \emph{sentence level cloze completion} with multiple sentence blanks and a shared candidate set with distractors. 
    \item We release \dsname, a sentence level cloze dataset of $\approx6$k passages and $\approx30$k blanks.
    \item We estimate human performance on \dsname, and benchmark several models, including state-of-the-art contextual embeddings (Table \ref{tab:apnres}). We find a significant gap of $>$ $15\%$ for future models to close in order to match human performance.
    \item  Through several ablations described in \S\ref{sec:discQuality}, we show that distractors designed by English teachers are of high quality and make the task more challenging. 
    \item We show that extra sentence level cloze questions generated automatically from an external corpus can be used to further improve model performance through data augmentation (See \S \ref{subsec:qa}).
\end{tight_enumerate}
\section{Related Work}
\label{sec:related}
\begin{table*}[t!]
    \centering
    \begin{tabular}{|l|l|l|l|l|l|l|}
    \toprule
    \textbf{Dataset} & SL & MB & Distractors & Candidates & Position  & $\|Context\|_{w}$    \\ 
    \midrule
    \textsc{\dsname} & \ding{51} & \ding{51} & Human & Shared & Anywhere  & 319   \\
    \midrule
    \textsc{ROCStories \shortcite{mostafazadeh2017lsdsem}} & \ding{51} & $\times$ & Human & - & End &  25 \\
    \textsc{CLOTH \shortcite{xie2017large}}  & $\times$ & \ding{51} & Human & Separated & Anywhere  & 243    \\ 
    \textsc{LAMBADA \shortcite{paperno2016lambada}} & $\times$ & $\times$ & Exhaustive & - & End &  76 \\ 
    \textsc{CBT \shortcite{hill2015goldilocks}}  & $\times$ & $\times$ & Automatic & - & End  & 465    \\ 
    \textsc{MRSCC \shortcite{zweig2011microsoft} }  & $\times$ & $\times$ & Human & -  & Anywhere & 20    \\ 
    \bottomrule
    \end{tabular}
    \caption{Comparing \dsname~with previous cloze datasets. Exhaustive denotes the case where the entire vocabulary is a candidate for a word level cloze. For the single-blank case, candidate sharing is irrelevant. SL and MB mean sentence level and multi-blanks respectively. $\|Context\|_{w}$ is the average token length of the context.}
    \textbf{\label{table:comparison}}
    \vspace{-5ex}
\end{table*}
Several cloze test datasets are collected to measure reading comprehension ability of machines. CNN/DailyMail \cite{hermann2015teaching}, an early  dataset of current QA research, constructs cloze questions from article summaries, with article spans as answers. Their cloze gaps are entities and hence one or few tokens long at best. The LAMBADA dataset \cite{paperno2016lambada} constructs a corpus of word level cloze gaps, such that each gap is in the last passage sentence. CBT \cite{hill2016automatic} creates word level cloze questions by removing a word in the last sentence of every consecutive 21 sentences, with the first 20 sentences being the context. \newcite{onishi2016did} curate a dataset of who-did-what type sentences with word level blanks. The CLOTH~\cite{xie2017large} dataset collects word level cloze questions from English exams designed by teachers. MRSCC \cite{zweig2011microsoft} consists of 1,040 word level cloze questions created by human annotators. 

Among recent cloze datasets, ROCStories \cite{mostafazadeh2017lsdsem} is the closest we could find to a sentence level cloze dataset. In this task, the first 4 sentences of a 5-sentence story are provided, and the task is to choose the correct ending from a pair of candidate ending sentences. However, there are several key differences between \dsname~and ROCStories. Firstly, there are multi-blanks in \dsname~which are not in a fixed position and require learning cues from bidirectional contexts of varying lengths. Secondly, the endings in ROCStories have been found to contain \emph{``annotation artifacts"} \cite{gururangan2018annotation} which makes a large fraction of them predictable independent of context. 

In contrast, \dsname~ is by design independent of artifacts, since
 a) given a blank, only some of our candidates are distractors, the rest being answers for other blanks. Even if one were to learn a classifier to distinguish distractors without context, the non-distractor candidates would be unresolvable without context. b) we further check how distinguishable our distractors are from non-distractors without context  by training a  strong classifier in this setting, as described in \S \ref{sec:discQuality}. The classifier obtains a reasonably low F1 score of $0.38$.

In Table \ref{table:comparison}, we summarize the comparison of \dsname~with cloze datasets from prior art to show its attractive aspects.

Public school examinations have been used as a data source by many earlier QA works, two prominent examples being the CLEF QA tracks \cite{penas2014overview,rodrigo2015overview} and RACE \cite{lai2017race}.
% \section{Why \textsc{AA} Could Hurt Generalization}
% \label{sec:motivation}
% \input{sections/motivation.tex}

%\section{Background}
%\label{sec:background}
%\input{sections/background.tex}

\section{\dsname~Dataset}
\label{sec:senta}
\subsection{Sentence Cloze Test with distractors}
In this task, each question consists of a passage, $S$, multiple sentence level  blanks $B$,  and  a  shared  set  of  candidates $C$ with distractors $D$, where $D \subset C$.
\paragraph{Problem Complexity\footnote{We defer the derivation to Appendix \S 1}} 

For our case, given the typical value of $|C|$ and $|B|$ being 7 and 5 respectively, the size of the answer space, $|A|$ is 2520.
Thus, the chance of guessing all blanks correctly at random is only 0.04\%. Moreover, there is a $48.2$\% probability of being entirely wrong with randomly guessing. Finally, given an answer list chosen uniformly at random, the expectation of number of distractors in the answer list is 1.4, i.e. on average, roughly one and half answers are distractors.

\subsection{Data Collection and Statistics} \label{sec:dataCollection}
Raw sentence cloze problems are crawled from public websites\footnote{http://www.21cnjy.com/;
http://5utk.ks5u.com/; http://zujuan.xkw.com/;
https://www.gzenxx.com/Html/rw/.} which curate middle and high school English exams designed by teachers. In total, 14,062 raw passages and 68,515 blank questions are crawled from these websites and the following steps are used to clean them. Firstly, duplicate passages are removed. Secondly, when the official answer to the problems are images, two OCR toolkits\footnote{\href{https://github.com/tesseract-ocr}{tesseract}; \href{https://www.abbyy.com/en-us/finereader/}{ABBYY FineReader}
} are employed to convert these images to text and the questions with different results from these two programs will be discarded. Finally, we remove examples which have 1) answers pointing to non-existent candidates, 2) missing or null candidates, 3) number of blanks $>$ number of candidates, 4) missing answers.

\begin{table}
    \centering
    \begin{tabular}{l l}
    \toprule
    \textbf{Statistic} & \textbf{Value}    \\ \midrule
    Total Passages & 5,959   \\
    Total Blanks & 29,731 \\
    Blanks Per Passage  & 4.99     \\ 
    \# Candidates Per Passage & 6.79  \\
    Avg Candidates Per Blank  & 1.35      \\ 
    \% Consecutive Blanks & 1.28  \\
    \# Words Per Passage & 319.64  \\
    Vocabulary Size & 48.6k\\
    {\rm Var}(Candidate Length)& 19.54 \\
    \bottomrule
    \end{tabular}
    \caption{\dsname~Statistics. For Consecutive Blanks, either of previous or next sentences is also a blank.}
    \textbf{\label{table:stats}}
    \vspace{-4ex}
\end{table} %

After cleaning, we obtain our \dsname~dataset with 5,959 passages and 29,731 blanks. They are randomly split into training, validation and test sets with 4790, 511 and 658 passages respectively.  The detailed statistics are presented in Table~\ref{table:stats}. We find that candidates have very different lengths and passages have long context.

\begin{table*}[!ht]
    \small
    \setlength{\tabcolsep}{4pt}
    \begin{tabular}{c l}
    \toprule %\hline
          \begin{tabular}{l} Type \end{tabular}  &   \begin{tabular}{l} Examples with Excerpts From Blank Context \end{tabular}    \\ \midrule
        
        \begin{tabular}{c} WM \\ (18.47\%) \end{tabular}  & \begin{tabular}{p{0.85\textwidth}} \textbf{1}: One day, a teacher was giving a speech to his student. He held up a \emph{glass of water} and asked the class  $\bf \rule{1cm}{0.15mm}$ The students answers ranged from 20g to 500g.
        
        \textbf{\ding{51} Candidate}:  B. How heavy do you think this \emph{glass of water} is?
        
        \textbf{$\times$ Candidate}: D. It does not matter on the weight itself.
        
        \textbf{Explanation}: WordMatch based on \emph{glass of water}.
         \end{tabular}   \\ 
         \midrule %\hline
        \begin{tabular}{c} Para. \\ (19.48\%) \end{tabular} & \begin{tabular}{p{0.85\textwidth}} \textbf{2}: If you want time to have breakfast with your family, save some time the night before by setting out clothes, shoes and bags. $\bf \rule{1cm}{0.15mm}$ That's a \emph{quarter-hour} more you could be sleeping if you bought a \emph{coffee} maker with a timer. 

        \textbf{$\times$ Candidate:} D. And consider setting a second alarm.
        
        \textbf{$\times$ Candidate}: F. Stick to your set bedtime and wake-up time, no matter the day.
        
        \textbf{\ding{51} Candidate}: G. Reconsider the \emph{15 minutes} you spend in line at the \emph{cafe}.
        
        \textbf{Explanation:} Need to match \emph{15 minutes}, \emph{quarter-hour} and \emph{coffee, cafe}. 
        \end{tabular}   
        \\ \midrule %\hline
        
        \begin{tabular}{c} Infer. \\ (41.97\%) \end{tabular}   & \begin{tabular}{p{0.85\textwidth}} \textbf{3}: May is a great month. $\bf \rule{1cm}{0.15mm}$ You can have a good time with your family.
                
        \textbf{$\times$ Candidate}: E. All the students can come to their schools.
        
        \textbf{\ding{51} Candidate}: F. From May 1st to 7th, we don't need to come to school.
        
        \textbf{$\times$ Candidate}: G. On May 20th, a famous sports star YaoMing comes to our school.

        \textbf{Explanation:} Need to infer that not coming to school $\rightarrow$ one is at home with family. Simply matching for words \emph{May} or \emph{school} will also match wrong candidates.
        \end{tabular}  \\ 
        \midrule %\hline
        \begin{tabular}{c} Sum. \\ (20.08\%) \end{tabular}   & \begin{tabular}{p{0.85\textwidth}}
        \textbf{4}: How to Enjoy Life As a Teen? Are high school days equal to the ``best years of your life''? Maybe not, but you can learn to make the most of your high school days $\bf \rule{1cm}{0.15mm}$ Whether it 's having a computer, having friends, having a good supply of food, a bed to sleep on, family that loves you, having a decent education or simply being born in this world. Be happy, and life will reward you.
        
        \textbf{$\times$ Candidate}: A. Remember that the point of life is for you to enjoy it.
        
        \textbf{\ding{51} Candidate}: C. Learn to appreciate small things.
        
        \textbf{Explanation}: After summarizing sentences after the blank [which describe a list of ``small things"], the answer should be C. A is a strong distractor since both ``enjoy'' and ``life'' appear in the context, besides being pertinent to the topic. Indeed, our best-performing BERT-ft model chooses A as the answer.
        \\ \end{tabular}   \\ \bottomrule %\hline
    \end{tabular}
    \caption{Blanks in a sample of 100 passages are manually categorized into four categories. For the ease of illustration, we've shown only limited  context around the blanks , and 1-2 wrong candidates. WM, Para., Infer. and Sum denote WordMatch, Paraphrase, Inference and Summary respectively. More examples are in Appendix.}
    \label{tab:categoriesTwo}
    \vspace{-2ex}
\end{table*}

\subsection{In-Depth Analysis \& Categorization}
\label{subsec:reasoningType}

In order to evaluate students’ mastery of a language, teachers usually design tests in a way that
questions cover different aspects of a language.

\paragraph{Reasoning Types}As illustrated with examples in Table \ref{tab:categoriesTwo}, we set a four-fold categorization for the reasoning which leads to a ground truth candidate being assigned to a blank. Our reasoning type taxonomy is motivated by categorization of question types in earlier works in QA such as \cite{chen2016thorough,trischler2017newsqa}\footnote{See Section 4.2 from both respective papers.}. Strictly speaking, these reasoning types could co-exist. But for simplicity, we classify each blank into only one of the four.
\begin{tight_itemize}
    \item \textsc{WordMatch}: If the candidate has word overlap, especially of non-stopwords or infrequent phrases, with context around the blank. 
    \item \textsc{Paraphrase}: If the candidate doesn't have an explicit word overlap with the context, but nevertheless contains words or phrases which are paraphrases of those in the context.
    \item \textsc{Inference}: If the candidate is a valid hypothesis conditioned on the left context [as premise], or a necessary pre-condition/premise based on the right context. Note that the candidate in this case doesn't contain word overlap/paraphrases which would obviate need for inferential reasoning. The reasoning required needs not be just strict entailment \cite{bowman2015large,marelli2014semeval} but could also involve abductive reasoning \cite{bhagavatula2019abductive}, where the candidate is just one of many likely hypothesis (premise) given the left (right) context as premise (hypothesis).
    \item \textsc{Summary}: 
    If the candidate is a summary, introduction, or conclusion of multiple sentences before or after it. 
    In this type, unlike  \textsc{Inference}, there is no requirement to deduce and reason about new hypotheses/possibilities not present in the premise - only consolidation and rearranging of information is required. 
\end{tight_itemize}
A sample of 100 passages containing 500 blanks are manually categorized into these four categories.
Examples and statistics of these four types are listed in Table~\ref{tab:categoriesTwo}. More than 40\% blanks need inference to be solved, denoting the high difficulty of our dataset.

\section{Methods}
\label{sec:methods}
\subsection{Context Length}
\label{sec:feature}
We experiment with giving our models different amounts of context. Through this, we can explore how context length affects model performance.
\vspace{-\topsep}
\begin{tight_enumerate}
  \setlength{\parskip}{0pt}
  \setlength{\itemsep}{0pt plus 1pt}
    \item \textsc{P(N)}: Immediate previous (next) sentence 
    \item \textsc{P+N}: Immediate previous and next sentence
    \item \textsc{AP(AN)}: All previous (next) sentences 
    \item \textsc{AP+AN}: All previous and next sentences
\end{tight_enumerate}
\vspace{-\topsep}
\textsc{AP+AN} is the unablated setting, where all passage sentences are available to the model.
%\subsection{Unsupervised Methods}
\subsection{PMI} %based on n-gram counting
Before exploring deep representational approaches, we would like to find how well symbolic ones perform at this task. Starting with works such as \newcite{iyyer2015deep} and \newcite{arora2016simple}, it has become convention to first benchmark simple baselines of this kind.  
PMI merely encodes how likely it is for a word pair to occur in consecutive sentences. It does not consider the internal sentence structures, or the relative position of the words in their respective sentence. Intuitively, it can be called a ``surface-level" approach. A high performance by PMI would indicate that candidates can be matched to blanks by simple ngram statistics, without requiring sentence representation, which would make \dsname~uninteresting.

We estimate PMI counts \cite{church1990word} from all consecutive sentence pairs in our training split. Let $\textrm{f}$ denote frequency
\begin{align*}
    \textrm{PMI}(w_{s},w_{c}) &= \frac{\textrm{f}(w_{s} \in S, w_{c} \in C)}{\textrm{f}(w_s \in S)\textrm{f}(w_c \in C)}
\end{align*}
Note that our PMI definition diverges from typical PMI since its asymmetric between $w_s$ and $w_c$. Since $S$ and $C$ are the sets of non-terminating and non-starting sentences respectively, they overlap but aren't identical. For a pair of sentences, we find aggregate $\overline{\textrm{PMI}}(S,C)$ as:
\begin{align*}
    \overline{\textrm{PMI}}(S,C) = \frac{1}{|C||S|} \sum_{w_{c} \in C} \sum_{w_{s} \in S} \textrm{PMI}(w_{s},w_{c})
\end{align*}
This definition can be extended to all n-grams upto a certain $n$. We denote this by $\overline{\textrm{PMI}}_{n}$. We notice that $\overline{\textrm{PMI}}_{n}$ performance saturates after $n=2$. Hence, in our experiments, we use $\overline{\textrm{PMI}}_{2}$. 

\subsection{Language Modelling}
One intuitive way to solve this task is to generate the blank sentence given the context by advanced pre-trained language models (LM). Formally, suppose the blank is the $i$th sentence, $s_{i}$, and $s_{1}, \dots, s_{i-1}$, $s_{i+1}, \dots, s_{n}$ are the context. Our goal is to choose $c_{k}$ from the candidate set which could maximize the joint probability $p(s_{1},\ldots,s_{i-1},c_{k},s_{i+1},\dots, s_{n})$.

Due to limited number of passages available to train a robust LM, Transformer-XL (TR.XL) Base~\cite{dai2019transformer}, trained on WikiText-103, is employed to address this task. In order to make decoding time tractable, context length is limited to three sentences before and after the blank.

\subsection{Coherence}
Coherence models assign a continuous score to a sentence sequence indicative of its coherence. This score is usually unnormalized and not needed to be a probability [unlike language models].

We use the local coherence approaches implemented by the \textsc{COHERE}\footnote{\url{github.com/karins/CoherenceFramework}} framework \cite{smith2016cohere}. Roughly, this model works on the intuition that successive sentences exhibit regularities in syntactic patterns. Specifically, it uses n-gram patterns on linearized syntactic parses (e.g. S NP VP \ldots) of consecutive sentences. Once trained, this model can return a ``coherence score" for any sentence sequence.

The \textsc{COHERE} model is first trained on all ground-truth passages from our training set, with the ground truth answers filled into the blanks. At test-time, we score each possible answer permutation using the trained \textsc{COHERE} model and pick the highest scoring one. Note that decoding for \textsc{COHERE} is by definition exhaustive, and doesn't make any assumptions by answering the blanks in a particular order.

\subsection{InferSent}
\newcite{conneau2017supervised} use textual inference supervision as a signal to train a shared sentence encoder for premises and hypotheses, which can later be used as a sentence representor. We refer to this approach as \textsc{INFST}.
Context features of a given blank and one candidate feed to two encoders in \textsc{INFST} respectively and classify whether this candidate is suitable to this blank. The maximum tokens of context features is set as 256. Bi-directional LSTMs with the max pooling operation are employed as our encoders. We follow the training procedure described in \newcite{conneau2017supervised}.
%\subsubsection{Rule-Based}
\subsection{BERT Models}
\paragraph{Input Representations} \label{subsubsec:bertRepn}

Let $c_{k}$ denotes the $k$th candidate. $s_{\minus i}$ and $s_{+i}$ denote the $i$th sentence before and after the blank respectively and $|P|$ and $|N|$ represent total number of sentences before and after the current blank respectively. Following the input convention in \newcite{devlin2018bert}, the input sequence given various context lengths and $c_{k}$ is:
\vspace{-\topsep}
\begin{tight_enumerate}
    \item \textsc{P} : $[\textrm{CLS}]s_{\minus1}[\textrm{SEP}]c_{k}$
    \item \textsc{N} : $[\textrm{CLS}]c_{k}[\textrm{SEP}]s_{+1}$
    \item \textsc{AP} : $[\textrm{CLS}]s_{\minus|P|}\ldots s_{\minus1}[\textrm{SEP}]c_{k}$
    \item \textsc{AN} : $[\textrm{CLS}]c_{k}[\textrm{SEP}]s_{+1}\ldots s_{+|N|}$
\end{tight_enumerate}

To retain sentence sequentiality, the order between the context and the candidate follows that in the original passage. Furthermore, for \textsc{(A)P+(A)N}, we create and score one input sample for each of the context directions during prediction. The average of these two scores is taken as the final score.
The maximum tokens of input is set as 256 in our experiments and only the context is truncated to meet this requirement.
\paragraph{BERT Next Sentence Prediction (NSP)}
One of the objectives in BERT pre-training stage is understanding the relationship between two sentences, which is highly correlated with our task. Therefore, we use the pre-trained BERT-Large-uncasedd  with its NSP layer to predict the most appropriate candidate for each blank given its context. Specifically, BERT is employed to predict the probability of the context and the candidate being consecutive. 
\paragraph{Finetuning BERT}
A wide range of NLP tasks have greatly benefited from the
pre-trained BERT model. Therefore, we also finetune the pre-trained BERT-Large model on our task through sequence pair classification schema. 
Specifically, for each blank, its correct candidate will be labelled as 0 and the label of all other wrong candidates is 1. Batch size and number of epochs for all models are 32 and 3. We employ Adam~\cite{kingma2014adam} as the optimizer with three different learning rates $\{1e^{-5}, 2e^{-5}, 3e^{-5}\}$. Best model selection is based on validation performance. All BERT finetuning experiments including ablation study follow this training strategy.

\section{Experiments}
\label{sec:experiments}
\begin{table}
\centering
\begin{tabular}{l l l}
\toprule
\bf Type & \bf Model & \bf{BA/PA}   \\ 
\midrule
\multirow{2}{*}{\textsc{UNSUP}} 
& \textsc{BERT} & 36.9/3.5   \\
  &\textsc{TR.XL} & 32.3/2.6 \\
 \midrule
\multirow{1}{*}{\textsc{FT}} & \textsc{BERT} &  \textbf{71.7}/\textbf{29.9}\\
 \midrule
  \multirow{3}{*}{\textsc{SUP}}
& \textsc{$\overline{\textrm{PMI}}_{2}$}  & 29.8/8.4    \\
%& \textsc{\textsc{Cohere}} & 25.7/5.4    \\
& \textsc{\textsc{Cohere}} & 23.3/1.1    \\
& \textsc{INFST} & 55.8/18.4\\
\midrule
\multirow{1}{*}{\textsc{Human}}  
& -& \textbf{87.1}/\textbf{56.3}    \\
\bottomrule
\end{tabular}
\caption{Test \textsc{BA}/\textsc{PA} of various model types with \textsc{Exh} decoding and AP+AN context.}
\label{tab:apnres}
% \vspace{-2ex}
\end{table}

\begin{table*}
\centering
\begin{tabular}{l r l l l l l l}
\toprule
\bf Type & \bf Model &  \textsc{P}  &  \textsc{N} & \textsc{AP}  & \textsc{AN} & \textsc{P+N}   & \textsc{AP+AN}   \\ 
\midrule
\multirow{2}{*}{\textsc{UNSUP}} 
& \textsc{BERT+\textsc{Inc}} & 33.0/2.1 & 34.7/4.1 & 29.8/2.1  & 15.7/0.3  & 34.7/2.3 & 27.3/1.4  \\
& \textsc{+\textsc{Exh}} & 34.2/3.2 & 40.2/4.7 & 31.5/2.6  & 14.7/0.0 & 40.2/4.7 & 36.9/3.5   \\
\midrule
\multirow{2}{*}{\textsc{FT}} & \textsc{BERT+\textsc{Inc}} & 44.3/6.8 & 48.0/9.6 & 50.4/9.9 & 56.9/16.1 & 61.0/20.4 & 66.6/25.1 \\
& \textsc{+\textsc{Exh}} & 47.2/8.5 & 54.2/11.2 & 60.0/17.5 & 60.0/17.5 & 66.5/25.2 & \textbf{71.7}/\textbf{29.9}\\
 \midrule
  \multirow{2}{*}{\textsc{SUP}}
& \textsc{$\overline{\textrm{PMI}}_{2}$+\textsc{Inc}} & 23.4/1.2  & 24.4/1.5 & 16.2/0.3 &  17.5/0.1 & 26.2/1.7 & 17.1/0.0   \\
& \textsc{+\textsc{Exh}} & 24.7/1.5  & 28.2/1.5 & 20.6/0.9 &  13.3/0.0 & 29.7/2.6 & 25.2/0.6  \\
\bottomrule
\end{tabular}
\caption{Test \textsc{BA}/\textsc{PA} of various model types unsupervised (\textsc{UNSUP}), finetuned (\textsc{FT}) and supervised (\textsc{SUP}) across varying context levels, with \textsc{Inc} or \textsc{Exh} decoding.}
\label{tab:mainres}
\end{table*}

\subsection{Decoding Strategy}
The decoding strategy decides how exactly we assign a candidate to each blank in the passage. Due to shared candidates, we have two strategies:
\begin{tight_enumerate}
  \setlength{\parskip}{0pt}
  \setlength{\itemsep}{0pt plus 1pt}
    \item \textsc{\textbf{Inc:}}  Answering each blank from left to right in order. Once a blank is answered with a candidate, this candidate is unavailable for later blanks. 
    \item \textsc{\textbf{Exh:}} Exhaustively scoring all permutations of candidates to answer the blanks. The score of a permutation is simply the sum of each its constituent blank-candidate pairs. The highest scoring permutation is the answer.
\end{tight_enumerate}

\subsection{Evaluation Metrics}
We design two metrics to evaluate models. Both of these metrics are reported as percentage.
\paragraph{Blank accuracy (BA):} The fraction of blanks answered correctly, averaged over all passages.
\paragraph{Passage Accuracy (PA):} PA is 1 \emph{iff} the model gets all blanks in a passage correct, and 0 otherwise. The average of PA over all passages is reported.

\begin{table}
    \centering
    \begin{tabular}{l|c|c|c}
    \toprule
         & BERT-Un & TR.XL & BERT-ft\\
         \midrule
        RemoveDt & 47.4/17.2 & 39.7/9.1  & 80.9/62.0\\
        RandomDt & 44.6/12.4 & 36.0/6.8  & 77.9/50.9\\
        Unablated & 40.2/4.7 & 32.3/2.6 & 71.7/29.9\\
        \bottomrule
    \end{tabular}
    \caption{Test BA/PA with distractor ablations on test set. RemoveDt and RandomDt represent removing and sampling distractors respectively. BERT-Un and BERT-ft denotes pre-trained and finetuned BERT.}
    \label{tab:ablationres}
\end{table}

\subsection{Human Performance}
We hire annotators from AMT to both answer and label difficulty for 144 randomly chosen test examples. Annotators are restricted to be from USA/UK and have \emph{Master} designation on AMT\footnote{Marked by AMT based on approval \%, no. approved etc.}, along with $>$ $90$\% HIT approval rate. 
On average, each annotator spends 624 seconds to answer one example. Difficulty level is chosen from \{\textit{VeryHard}, \textit{Hard}, \textit{Moderate}, \textit{Easy}, \textit{VeryEasy}\}. $3.5$\% of annotators find the task \textit{VeryHard}, while $8.3\%$ find it \textit{VeryEasy}. The largest fraction of $38.2\%$ find it to be \textit{Moderate}. We note that \dsname~contains a larger proportion of non-easy questions (61.0\%). Human performance is reported in Table~\ref{tab:apnres}. Annotators achieve BA of 87\%
which we take as the ceiling performance for models to match. 

\subsection{Model Performance}
All models are trained with AP+AN context and decoded by \textsc{Exh}\footnote{Unless stated otherwise, models decode with EXH and are trained with full context i.e AP+AN}. Results are shown in Table~\ref{tab:apnres}. Finetuning BERT achieves the best performance among other models, though it still lags behind human performance significantly. Unsupervised models could only solve one third of all blanks. Surprisingly, $\overline{\textrm{PMI}}_{2}$ and \textsc{COHERE} performs worse than the unsupervised models.  We conjecture that it is difficult for COHERE, using syntactic regularities alone, to distinguish between the ground truth answer for a particular blank and another candidate which is a ground truth answer for another nearby blank. As noted, $\overline{\textrm{PMI}}_{2}$ suffers due to inability of incorporating larger context.

\label{sec:model_experiments}
% The main result of all models is shown in Table~\ref{tab:mainres}. 
To explore effects of various context length and decoding strategies, models are trained with different context lengths and inferred by both decoding methods. Results are shown in Table~\ref{tab:mainres}.
\paragraph{\textsc{Inc} vs \textsc{Exh}} \textsc{Exh} is better than \textsc{Inc} for most approaches, indicating that human created blanks are interdependent and need joint answering. 
\paragraph{Context Length} Increasing the context length, such as (P vs. AP), could significantly improve model performance, showing that this task needs discourse-level context to be answered. Furthermore, models with bidirectional context, such as (P+N), perform better than single-direction context, e.g., P, indicating that this task needs global context. Lastly, we observe that PMI-based approaches which do not explicitly encode sentences are unable to incorporate larger context levels, showing best performance with \textsc{P+N}.

\subsection{BERT-ft vs. Human}

\begin{figure}
    \centering
    \includegraphics[width=0.9\columnwidth]{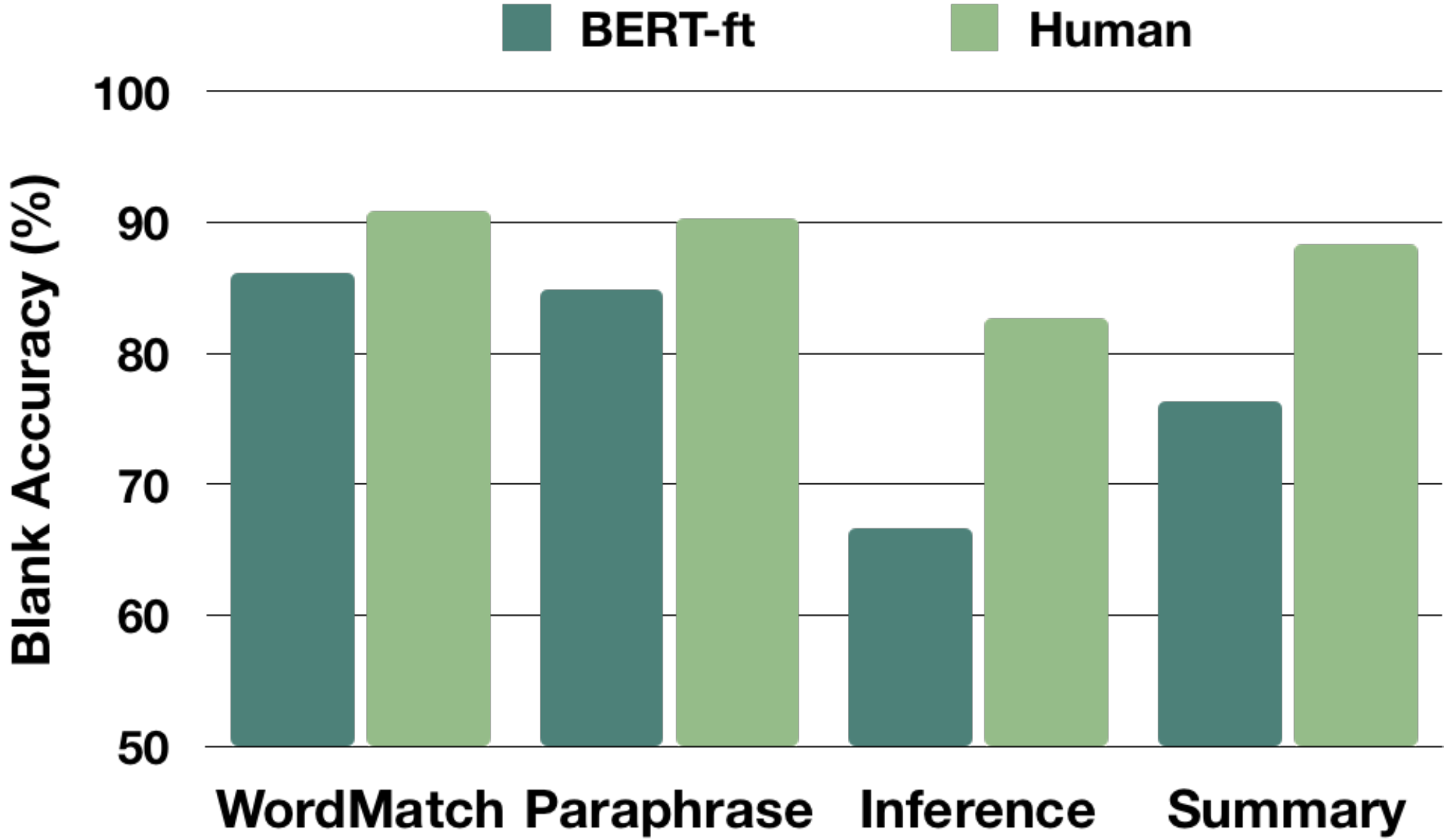}
    \caption{Test blank accuracy of BERT-ft and Human on each reasoning type category introduced in \S \ref{subsec:reasoningType}.}
    \label{fig:error}
    % \vspace{-1ex}
\end{figure}

BERT after finetuning (BERT-ft) can perform reasonably well (72\%) but there is still a gap comparing with human performance (87\%). In this section, we would like to analyze the strength and weakness of BERT-ft compared with \textsc{Human}. Therefore,
we analyze their performance across different reasoning categories on test set. From Figure~\ref{fig:error}, inference questions are the most difficult for both \textsc{Human} and BERT-ft and questions needing WordMatch are relatively easy.
Compared with human performance, BERT-ft could achieve comparable BA on WordMatch and paraphrasing problems. However, BERT-ft performs much worse on questions needing inference and summary. 
We also refer to some examples from Table \ref{tab:categoriesTwo}. 

In \emph{Example $4$}, BERT-ft prefers A but the answer is C. The reason why BERT-ft chooses A may be that ``enjoy life'' happens in the context, but summarizing the next sentence is necessary to achieve the correct answer. Therefore, it is necessary to improve the ability of BERT to represent meaning at the sentence level beyond representing individual words in context.  

We also explore how the system performance corresponds to the human judgement of difficulty. Since evaluates rate the problems into 5 difficulty levels, we report the system BA/PA for each level in Table~\ref{tab:diffculty_corr}. For BA (blank-level accuracy), we see that, overall, the system accuracy decreases as difficulty increases from VeryEasy (0.75) to VeryHard (0.68). However, the decrease is not exactly monotonic (there is a small increase from VeryEasy to Easy, as also from Moderate to Hard). 

We conjecture that non-monotonicity could be due to two reasons:
\begin{itemize}
    \item Our difficulty annotations are at passage level rather than blank level. There might be some hard blanks in a passage marked overall “Easy”. Conversely, there might be easy blanks in a passage marked overall “Hard”. 
    \item Since we’ve more examples marked with certain difficulty levels - e.g 30.5\% examples are “Easy” while only 8.3\% are “VeryEasy”. This might make system accuracy average for levels with more examples more stable (lower sample variance), leading to some non-monotonicity (e.g for Easy and VeryEasy)
\end{itemize}
For PA (passage-level accuracy, i.e., getting all questions correct) also, we see a clear decrease as difficulty increases from VeryEasy (0.63) to VeryHard(0.2). The decrease here is sharper than BA , with only one violation of monotonicity (increase from 0.29 to 0.35 on Moderate to Hard). The sharper trend for PA supports our first point above.

\begin{table}[!ht]
    \centering
    \begin{tabular}{l|c|c}
        \toprule
        Diffculty &  BA & PA\\
        \midrule
        Very Easy & 0.75 & 0.63 \\
        Easy & 0.78 & 0.45\\
        Moderate & 0.71 & 0.29\\
        Hard & 0.72 & 0.35\\
        Very Hard & 0.68 & 0.20\\
        \bottomrule
    \end{tabular}
    \caption{BERT-ft performance in terms of human judgement of diffculty.}
    \label{tab:diffculty_corr}
\end{table}
\subsection{Distractor Quality} \label{sec:discQuality}
An attractive aspect of this task is distractors designed by English teachers. We verify distractor quality through the following experiments.
\paragraph{Model Performance w/o Distractors} All distractors in the test set are removed and models are evaluated on this non-distracted test set. Results are shown in Table~\ref{tab:ablationres}. It is clear to see that after removing these distracting candidates, models can get better scores, showing that models find it hard to exclude distractors during prediction.
\paragraph{Randomly Sampled Distractors} After removing human-created distractors, we further randomly sample sentences from other passages as new distractors. To mitigate sampling variance, we run this experiment with 8 seeds and report the averaged score in Table~\ref{tab:ablationres}. Comparing with distractors designed by teachers, models could discern these distractors more easily.
\paragraph{Annotation artifacts of distractors} Annotation artifacts~\cite{gururangan2018annotation} occurs in many datasets created by human annotators. A potential artifact type for our task is whether we could detect distractors without passages.  Therefore, we finetune BERT-Large as a binary classifier, the input of which is just distractors and other correct candidates. With this model, we could only obtain 38\% F1 score on the test set, showing that it is difficult to filter distractors out without any context.

\paragraph{Distractor Error (DE)}  We define DE as the number of predicted answers per passage which are actually distractors. Through DE, we measure a model's ability to exclude distractors during prediction. Results are shown in Table~\ref{tab:de_res}. \textsc{Human} has the lowest DE and BERT-ft could discern distractors to some extent. However, DE of $\overline{\textrm{PMI}}_{2}$  is more than 1, meaning that on average, there is atleast one distractor in the predicted answer list.

\begin{table}
    \centering
    \begin{tabular}{c|cccc}
    \toprule
        Model & Uni. &  $\overline{\textrm{PMI}}_{2}$  & BERT-ft & \textsc{Human} \\
        \midrule
        DE & 1.429 & 1.204 & 0.661 & \textbf{0.375}  \\
        \bottomrule
    \end{tabular}
    \caption{Distractor error on test set of different models. Uni. denotes the uniform model.}
    \label{tab:de_res}
    % \vspace{-3ex}
\end{table}

\begin{table}
    \centering
    \begin{tabular}{l|ccc}
    \toprule
        Training Strategy &  PA&BA&DE\\
        \midrule
        $Q_{A}$ &   65.2&26.1&0.792\\
        $Q_{H}$ &  71.7 & 29.9&0.661\\
        $Q_{A}$ ; $Q_{H}$ & 74.2&33.9&\bf0.624\\
        $Q_{A}$ + $Q_{H}$ & \bf74.5 & \bf34.3 & 0.637\\
        \bottomrule
    \end{tabular}
    \caption{Test performance of models with $Q_{A}$ and $Q_{H}$.}
    \label{tab:random_res}
    % % \vspace{-3ex}
\end{table}

In summary, distractors created by teachers are high quality and increase task difficulty. 
\subsection{Automatically Generated Sentence Cloze Questions}
\label{subsec:qa}
To explore automatic generation of examples for the task, we construct sentence cloze questions by randomly choosing five sentences in a passage as blanks. We defer automatically generating distractors to future work since non-trivial distractor generation is a hard problem in itself. Specifically, we extract all passages from RACE~\cite{lai2017race} (which is also from exams) and filter out passages which have less than 10 sentences or more than 30 sentences. While choosing blank positions, we prevent three or more blanks consecutive to each other in generated questions. Finally, 16,706 examples are obtained automatically. Here, questions generated automatically and collected from examinations are called $Q_{A}$ and $Q_{H}$ respectively.

We leverage $Q_{A}$ in three ways: 1). train models only on $Q_{A}$ , 2) first train models on $Q_{A}$ and finetune models on $Q_{H}$, i.e., $Q_{A}$ ; $Q_{H}$, 3) train models on the concatenation of $Q_{A}$ and $Q_{H}$, i.e., $Q_{A}$ + $Q_{H}$. BERT-Large is finetuned through these ways and results are shown in Table~\ref{tab:random_res}. The model trained only on $Q_{A}$ has worst performance and we attribute this to the difficulty of distinguishing distractors without seeing them during training. Therefore, this model has the highest DE. However, models trained on $Q_{H}$ and $Q_{A}$ could achieve better performance. We conjecture this is because $Q_{A}$ assists the model to have better generalization.

\section{Conclusion}
\label{sec:conclusions}
We introduce \dsname, a sentence cloze dataset with high quality distractors carefully designed by English teachers. \dsname~requires use of discourse-level context and different reasoning types. More importantly, the high quality distractors make this task more challenging. Human performance is found to exceed advanced contextual embedding and language models by a significant margin. Through \dsname, we aim to encourage the development of more advanced language understanding models.

\section{Acknowledgements}
We thank Qizhe Xie, Hiroaki Hayashi and the 3 anonymous reviewers for valuable comments.
\newpage
\bibliography{acl2019}
\bibliographystyle{acl_natbib}
\clearpage
\newpage
\section{Problem Complexity}
With $|B|=5$ blanks and $|C|=7$ candidates, the size of answer space, $|A|$, is number of permutations $|B|$ objects taken $|C|$ at a time, i.e., $\textrm{P}(7,5)=2520$. Therefore, the probability of answering all blanks correctly is $\frac{1}{2520}=0.03\%$

What are the chances of getting answers partially correct? What are the chances of getting answers partially correct? If we have the same number of candidates as blanks, this is equivalent to $|B|! - D_{|B|}$, where $D_{|B|}$ is the number of derangements\footnote{\url{en.wikipedia.org/wiki/Derangement}} of $|B|$ elements. In the presence of more candidates than blanks i.e distractors, this expression becomes more involved to derive. Therefore, here, we enumerate all the permutation of answer lists given a correct answer. With $|C|=7$ and $|B|=5$, $\zeta(|C|,|B|)=51.8\%$. In other words, there is a $48.2$\% probability of being entirely wrong with a randomly chosen set of answers to each blank in the passage.

What are the chances of getting distractors as predicted answers? For the expectation of number of distractors choosing by uniform model, it should be $\mathbb{E}[DE]$, where $DE$ denotes distractors errors.
\begin{equation}
    \sum_{d=0}^{2} p(DE=d)\times d
\end{equation}
where $p(DE=d)$ denotes the probability of d predicated answers are distractors. Since there are two distractors in candidates, the maximum of $d$ is 2.
Furthermore, $p(DE=1)$ is 
\begin{equation}
  \textrm{P}(5,4)\textrm{C}(5,4)\textrm{C}(2,1)/|A|=0.476
\end{equation}
and $p(DE=2)$ is
\begin{equation}
  \textrm{P}(5,3)\textrm{C}(5,3)\textrm{A}(2,2)/|A|=0.476
\end{equation}
where $\textrm{C}(\cdot,\cdot)$ and $\textrm{P}(\cdot,\cdot)$ is combination and permutation respectively. Therefore, the expectation of number of distractors is 1.429.

\begin{table*}[ht!]
    %\centering
    \captionsetup{font=footnotesize}
    %\captionsetup{font=small}
    \small
    %\tiny
    \setlength{\tabcolsep}{4pt}
    \begin{tabular}{l l}
    \toprule %\hline
          \begin{tabular}{l} Reasoning \end{tabular}  &   \begin{tabular}{l} Examples with Excerpts From Blank Context \end{tabular}    \\ \midrule
        
        \begin{tabular}{c} WM \\ (18.47\%) \end{tabular}  & \begin{tabular}{p{0.9\textwidth}} \textbf{1}: One day, a teacher was giving a speech to his student. He held up a \emph{glass of water} and asked the class.  $\bf \rule{1cm}{0.15mm}$ The students answers ranged from 20g to 500g.
        
        \textbf{\ding{51} Candidate}:  B. How heavy do you think this \emph{glass of water} is?
        
        \textbf{$\times$ Candidate}: D. It does not matter on the weight itself.
        
        \textbf{Explanation}: Match based on \emph{glass of water}\\\\

        \textbf{2}: Begin the sleep adjustment for your school schedule as \emph{early} as possible.
        
        $\bf \rule{1cm}{0.15mm}$  But if you feel you will need some extra time to adjust, \emph{start earlier}.
        
        \textbf{\ding{51} Candidate}: C. \emph{Starting} a few days \emph{early} will be enough.
        
        \textbf{$\times$ Candidate}: A. Relax before you go to bed.
        
        \textbf{Explanation}: Match based on \emph{early}, \emph{start} 
        \\ \end{tabular}\\
        \midrule %\hline
        \begin{tabular}{c} Para. \\ (19.47\%) \end{tabular} & \begin{tabular}{p{0.9\textwidth}} \textbf{3}: If you want time to have breakfast with your family, save some time the night before by setting out clothes, shoes, and bags. $\bf \rule{1cm}{0.15mm}$ That's a \emph{quarter-hour} more you could be sleeping if you bought a \emph{coffee} maker with a timer.
        
        \textbf{\ding{51} Candidate}: G. Reconsider the \emph{15 minutes} you spend in line at the \emph{cafe}. 
        
        \textbf{$\times$ Candidate}: F. Stick to your set bedtime and wake-up time, no matter the day.
        
        \textbf{$\times$ Candidate:} D. And consider setting a second alarm \\ \textbf{Explanation:} Need to match \emph{15 minutes}, \emph{quarter-hour} and \emph{coffee, cafe}  \\ \\
        
        \textbf{4}: Riding a London subway, a person from China will notice one major difference: In London,  commuters do not look at each other.  $\bf \rule{1cm}{0.15mm}$ That's not rudeness- people are just too busy to bother \emph{looking}.
        
        \textbf{\ding{51} Candidate}: E. In fact, \emph{eye contact} is avoided at all times.
        
        \textbf{$\times$ Candidate:} F. Apple must earn a fortune from London commuters. 
        
        \textbf{$\times$ Candidate:} G. Modern Londoner are fancy victims.
        
        \textbf{Explanation:} Need to match \emph{looking} and \emph{eye contact}
        \end{tabular}   \\ \midrule %\hline
        
        \begin{tabular}{c} Infer. \\ (41.16\%) \end{tabular}   & \begin{tabular}{p{0.9\textwidth}}  \textbf{5}: May is a great month. $\bf \rule{1cm}{0.15mm}$ You can have a good time with your family.
        
        \textbf{\ding{51} Candidate}: F. From May 1st to 7th, we don't need to come to school.
        
        \textbf{$\times$ Candidate}: G. On May 20th, a famous sports star YaoMing comes to our school.
        
        \textbf{$\times$ Candidate}: E. All the students can come to their schools.
        
        \textbf{Explanation:} Need to infer that not coming to school $\rightarrow$ one is at home with family. Simply matching for words \emph{May} or \emph{school} will also match wrong candidates. \\  \\ 
        
        \textbf{6}: The Colosseum in Rome was built during the time of the Roman Empire, in the first century AD. $\bf \rule{1cm}{0.15mm}$.  It is a popular tourist attraction today.
        
        \textbf{\ding{51} Candidate}: D. It could seat 50K people, who went to see fights between animals and people. 
        
        \textbf{$\times$ Candidate:} B. The country used to depend on agriculture.
        
        \textbf{$\times$ Candidate:} C. Mountains cover about three-fourths of the country.
        
        \textbf{Explanation:} World knowledge that \emph{Colosseum} or \emph{-eum} suffix relates to building with seating facility. Also coreference with the \textit{It} in \emph{It is a popular \ldots}  \\ \\ 
        \textbf{7}: American students usually get to school at about 8 : 30 in the morning. $\bf \rule{1cm}{0.15mm}$ In class, American students can sit in their seats when they answer teachers' questions.
        
        \textbf{\ding{51} Candidate}: B. School starts at 9:00 a.m.
        
        \textbf{$\times$ Candidate:} D. Then they take part in different kinds of after-school activities.
        
        \textbf{Explanation:} Requires inference about time. Activity starts at 9 after  participants get there before. \\ \end{tabular}  \\ \midrule %\hline
        
        \begin{tabular}{c} Sum. \\ (20.08\%) \end{tabular}   & \begin{tabular}{p{0.9\textwidth}} 
        \textbf{8}: Around water, adults should watch children at all times to make sure they are safe. Those who don't know how to swim should wear life jackets.  But by themselves they are not enough, so an adult should always  be present.  If you have to rescue a child from drowning, a few seconds can make a big difference. Make sure you have  a friend with you whenever you swim. $\bf \rule{1cm}{0.15mm}$. That person can make sure you get help. Drink a lot water. The sun's heat and the physical activity may make you sweat  more  than you realize. By following these simple rules, you can make sure your swim time is safe as well as fun. $\bf \rule{1cm}{0.15mm}$
        
        \textbf{\ding{51} Candidate}: B. Now get out there, and enjoy the water.

        \textbf{$\times$ Candidate}: D. Make sure everyone in your family swim well.
        
        \textbf{Explanation}: B is a good conclusion pertinent to the content of the passage.
        \\
        \\
        \textbf{9}: $\bf \rule{1cm}{0.15mm}$. Whenever you are worried, write down the questions that make you worry. And write out all the various steps you could take and then the probable consequences of each step. For example, "What am l worrying about?", What can I do about it? Here is what I'm going to do about it. After carefully weighing all the facts, you can calmly come to a decision.

        \textbf{\ding{51} Candidate}: A. Analyze the facts.

        \textbf{$\times$ Candidate}: C. Decide how much anxiety a thing may be worth.

        \textbf{Explanation}: A is a more appropriate option to summarize its succeeding context.
        \\
        \\
        \textbf{10}: Expect to work, $\bf \rule{1cm}{0.15mm}$. If you are not working you are not learning. You are wasting your time at school. Teachers can not make everything enjoyable.
        
        \textbf{\ding{51} Candidate}: F. School is not a holiday camp.
        
        \textbf{$\times$ Candidate}: E. Because it means that you are enjoying school and learning more.
        
        \textbf{Explanation}: After summarizing the sentences after the blank, the blank should be filled by F.
        \\ \end{tabular}   \\ \bottomrule %\hline
    \end{tabular}
    \caption{More examples of reasoning categories.}
    \label{tab:appendx_reason}
\end{table*}

\begin{table*}[ht]
    \small
    \begin{tabular}{|c|}
    \specialrule{\cmidrulewidth}{0pt}{0pt} 
         \begin{tabular}{p{0.98\textwidth}}
        Dear David ${\bf \rule{1cm}{0.15mm}}_{1}$ After I had spent a week with my English \textcolor{magenta}{family},  I slowly began to \textcolor{blue}{understand} their English a little better. ${\bf \rule{1cm}{0.15mm}}_{2}$ Students in my group are from different cities of  Britain and their \textcolor{blue}{dialects are different} too! Some of their \textcolor{blue}{accents} are quite strong  and they also have their own \textcolor{blue}{words and expressions}. ${\bf \rule{1cm}{0.15mm}}_{3}$ Before I came to England  I had thought that fish and chips were eaten every day. That's quite wrong! I get rather annoyed now when I hear all the foolish words about typical English food. I had expected to see ``London fog''. Do you remember our texts about it ? We had no idea that most of this ``thick fog'' disappeared many years ago when people stopped using coal in their homes. But the idea to speak about weather was very helpful. ${\bf \rule{1cm}{0.15mm}}_{4}$ On the other hand , habits are different . People tell me what is typical British here in London is not always typical in Wales or Scotland. ${\bf \rule{1cm}{0.15mm}}_{5}$ But what is ordinary for all British is that they follow traditions. Probably Britain has more living signs of its past than many other countries. And people have always been proud of having ancient buildings in capitals, big cities and the countryside. I will tell you more about Britain in my other letters. Love from Britain.
         \\ \\ 
         
         \textbf{Candidates:}

        A. But it's not the language that's different and surprising.

        B. Thanks for your nice letter.

        C. I have \textcolor{blue}{difficulty in understanding} my classmates.

        D. The \textcolor{magenta}{family} I live with are \textcolor{magenta}{friendly}.

        E. It 's very different from what I learned at school.

        F. Local habits and traditions are not the same as what we knew.

        G. The weather in London is really changeable.

        \\\\
        \textbf{Answers}: 1$\rightarrow$B , 2$\rightarrow$E, 3$\rightarrow$A , 4$\rightarrow$G, 5$\rightarrow$F (C and D are distractors)
        \\ \\
        \textbf{Discussion}: C is a strong distractor - not only does it have strong word overlap with the contexts of many blanks - it also has words which can make it rank high in terms of the possible inferences (\textcolor{blue}{dialects are different} implies \textcolor{blue}{difficulty in understanding}. Though not as strong as C, D also has a key word matching and is  similar in content to the topic.
        \\
        \end{tabular}   \\ \specialrule{\cmidrulewidth}{0pt}{0pt} %\hline
        \\
                 \begin{tabular}{p{0.98\textwidth}} 
        How to Enjoy Life As a Teen. Are high school days equal to the ``best years of your life''? Maybe not, but you can learn to make the most of your high school days. ${\bf \rule{1cm}{0.15mm}}_{1}$ Whether it's having a computer, having friends, having a good supply of food, a bed to sleep on, family that loves you, having a decent education or simply being born in this world. Be happy, and life will reward you. Remember that these are the last few years you will be able to enjoy yourself without having to worry about the responsibility of an adult, but make sure you prepare yourself for when you do become one. Choose your friends wisely. Unlike what many articles state, you don't have to be popular and have a gazillion friends to be happy. ${\bf \rule{1cm}{0.15mm}}_{2}$ Try to have friends that like you who you are, not just because you are wearing a certain brand of shoes or something like that. These are people who shop at the same store as you; not someone who will sympathize with you when your dog dies. ${\bf \rule{1cm}{0.15mm}}_{3}$ Participating in clubs, activities, and sports increases your chances of meeting new friends. While you only need 4 or 5 close friends, that doesn't mean you shouldn't try to meet new people. Participating gives you something to do instead of sitting bored at home and wallowing in self-pity. You can pursue interests you enjoy. Video games, for example, are good if you're the type who can get into that kind of thing. Use your ``\textcolor{magenta}{hobby} time'' either to gain \textcolor{magenta}{practical} skills for college apps, job resumes, and scholarships or get into something else in the creative field like painting or dance. ${\bf \rule{1cm}{0.15mm}}_{4}$ Work at a job you can \textcolor{blue}{enjoy}. Working is a great way to gain experience and to meet other people. When you do get out of college, interviewing companies will look at your prior work experience. ${\bf \rule{1cm}{0.15mm}}_{5}$ If you can't find work, especially in this hard economic time, volunteer or make your own job.
         \\ \\ \textbf{Candidates:}

        A.Remember that the point of life is for you to \textcolor{blue}{enjoy} it.

        B. In fact, many of the ``friends'' you have when you are popular are not true friends.

        C. Learn to appreciate small things.

        D. Be sociable.

        E. This will look great on your resume.

        F. This is the time to start developing passions.

        G. You should also find a \textcolor{magenta}{hobby} that is \textcolor{magenta}{meaningful} or practical.

        \\\\
        \textbf{Answers}: 1$\rightarrow$C , 2$\rightarrow$B, 3$\rightarrow$D , 4$\rightarrow$F, 5$\rightarrow$E (A and G are distractors)
        \\ \\
        \textbf{Discussion}: Both A and G are strong distractors especially for ${\bf \rule{1cm}{0.15mm}}_{4}$. Both of them overlap on key words, and do fit in the local context, though they are less coherent w.r.t F (which doesn't have any overlapping words) when placed in the broader narrative.
        \\
        \end{tabular}   \\ \specialrule{\cmidrulewidth}{0pt}{0pt} %\hline
        
    \end{tabular}
    \caption{Examples with strong distractors}
    \label{tab:strongDistractorExamples}
\end{table*}

\begin{table*}[ht]
    \captionsetup{font=footnotesize}
    \small
    \setlength{\tabcolsep}{4pt}
    \begin{tabular}{|l|}
    \specialrule{\cmidrulewidth}{0pt}{0pt}
         \begin{tabular}{p{0.98\textwidth}}
        The demand for ways to improve memory is higher in students than it is in adults. Students often come across new knowledge in  different areas that they need to store for exams. ${\bf \rule{1cm}{0.15mm}}_{1}$ Here are three effective \textcolor{blue}{ways} to \textcolor{magenta}{improve your memory} as a student. ${\bf \rule{1cm}{0.15mm}}_{2}$ Research shows that learning activities that take more than two hours without a break are less productive when compared to those that take one hour or 30 minutes. Students are likely to remember things they learn over a short period of time.  Make sure you take breaks between learning sessions to help improve your memory. Try to relax. Relaxing should be an essential part of your learning process. Scientists have proven that stronger and lasting memories can be achieved when a person \textcolor{orange}{relaxes}. ${\bf \rule{1cm}{0.15mm}}_{3}$ Deep breathing  is one of the most popular \textcolor{orange}{relaxation} \textcolor{blue}{techniques}. Establish a quiet environment and find a comfortable position. Then go through a deep breathing process for at least 15 minutes. Train the brain Students should give their brains a workout in order to improve their memory. At times the brain needs the right stimulation to keep growing and developing. You need to come up with a brain boosting activity that is suitable for you. ${\bf \rule{1cm}{0.15mm}}_{4}$ Write a short story and then try to use seven to nine words to describe it. You can also do games and puzzles to help \textcolor{magenta}{improve your memory}. \textcolor{teal}{${\bf \rule{1cm}{0.15mm}}_{5}$} The \textcolor{blue}{techniques} discussed above will help you to \textcolor{magenta}{improve your memory} significantly.
         \\ \\ \textbf{Candidates:} \\ 
        A. Distribute learning.\\
        B. Enrich learning activities.  \\ 
        C. Some students suffer with memory problems. \\
        D. Like a muscle memory can stretch and grow with a workout. \\
        E. For instance you can prepare a list of items and try to memorize them. \\
        F. You need to use different \textcolor{orange}{relaxation} \textcolor{blue}{techniques} in order to \textcolor{magenta}{improve your memory}. \\
        G. \textcolor{teal}{In summary} a good memory is an important advantage to any student who wants to improve his or her grades.  \\
        \\
        \textbf{Answers}: 1$\rightarrow$C, 2$\rightarrow$A, 3$\rightarrow$F , 4$\rightarrow$E, 5$\rightarrow$G (B and D are distractors)
        \\ \\
        \textbf{Discussion}: The candidate F can actually go into three possible blanks and fit well into their context - Blanks 1, 3 and 5. This can be seen from the several overlapping phrases/paraphrases F shares with all three, as shown by the three colors (one per concept). However, G (which starts with the phrase \emph{In summary,} can only fit into Blank 5. A is also difficult  to place in any blank other than Blank 1. Hence , candidate F has to be placed into Blank 3.
        \\
        \end{tabular}   \\ \specialrule{\cmidrulewidth}{0pt}{0pt} %\hline
        
    \end{tabular}
    \caption{Examples which require multi-blank logic}
    \label{tab:multiBlankLogicExamples}
\end{table*}

\begin{table*}[ht]
    \captionsetup{font=footnotesize}
    \small
    \setlength{\tabcolsep}{4pt}
    \begin{tabular}{|c|}
    \specialrule{\cmidrulewidth}{0pt}{0pt}
         \begin{tabular}{p{0.98\textwidth}}
        A student's life is never easy. And it is even more difficult if you will have to complete your study in a foreign land. ${\bf \rule{1cm}{0.15mm}}_{1}$ The following are some basic things you need to do before even seizing that passport and boarding on the plane. Knowing the country. You shouldn't bother researching the country's hottest tourist spots or historical places. You won't go there as a tourist, but as a student. ${\bf \rule{1cm}{0.15mm}}_{2}$ In addition, read about their laws. You surely don't want to face legal problems, especially if you're away from home. ${\bf \rule{1cm}{0.15mm}}_{3}$ Don't expect that you can graduate abroad without knowing even the basics of the language. Before leaving your home country, take online lessons to at least master some of their words and sentences. This will be useful in living and studying there. Doing this will also prepare you in communicating with those who can't speak English. Preparing for other needs. Check the conversion of your money to their local currency. ${\bf \rule{1cm}{0.15mm}}_{4}$ The Internet of your intended school will be very helpful in findings an apartment and helping you understand local currency. Remember, you're not only carrying your own reputation but your country's reputation as well. If you act foolishly, people there might think that all of your countrymen are foolish as well. ${\bf \rule{1cm}{0.15mm}}_{5}$ 
         \\ \\ \textbf{Candidates:} \\ 
        A. Studying their language. \\
        B. That would surely be a very bad start for your study abroad program.  \\ 
        C. Going with their trends will keep it from being too obvious that you're a foreigner. \\
        D. Set up your bank account so you can use it there , get an insurance , and find an apartment. \\
        E. It'll be helpful to read the most important points in their history and to read up on their culture. \\
        F. A lot of preparations are needed so you can be sure to go back home with a diploma and a bright future waiting for you. \\
        G. Packing your clothes.  \\
        \\
        \textbf{Answers with Reasoning Type}:\\ 1$\rightarrow$F (\textit{Summary}), 2$\rightarrow$E (\textit{Inference}), 3$\rightarrow$A (\textit{Paraphrase}), 4$\rightarrow$D (\textit{WordMatch}), 5$\rightarrow$B (\textit{Inference}) (C and G are distractors)
        \\ \\
        \textbf{Discussion}: Blank 3 is the easiest to solve, since \emph{Studying their language} is a near-paraphrase of \emph{Knowing even}  \emph{the basics of the language}. Blank 2 needs to be reasoned out by \textit{Inference} - specifically $E$ can be inferred from the previous sentence. Note however that $C$ is also a possible inference from the previous sentence - it is only after reading the entire context, which seems to be about learning various aspects of a country, that $E$ seems to fit better. Blank 1 needs to be reasoned out by \textit{Summary} $\rightarrow$ it requires understanding several later sentences and abstracting out that they all refer to \emph{lots of preparations}. Finally, Blank 5 can be mapped to B by inferring that \textit{people thinking all your countrymen are foolish} is \emph{bad}, while Blank 4 is a easy \textit{WordMatch} on \textit{apartment} to D.        
        \\
        \end{tabular}   \\ \specialrule{\cmidrulewidth}{0pt}{0pt} %\hline

        \begin{tabular}{p{0.98\textwidth}} 
        Latest news and comment on Street art from guardian.co.uk... ${\bf \rule{1cm}{0.15mm}}_{1}$  You can find it on buildings sidewalks street signs and trash cans from Tokyo to Paris from Moscow to Cape Town. Street art has become a global culture and even art museums and galleries are collecting the works of street artist. Street art started out very secretly because it was illegal to paint on public and private property without permission. ${\bf \rule{1cm}{0.15mm}}_{2}$ Some think it is a crime and others think it is a very beautiful new form of culture. Art experts claim that the street art movement began in New York in the 1960s.  Young adults painted words and other images on the walls and trains. This colorful style of writing became known as graffiti whose art showed that young people wanted to rebel against society. Street artists do their work for different reasons. ${\bf \rule{1cm}{0.15mm}}_{3}$ They choose street art because it is closer to the people. Some artists try to express their political opinion in their work. Others like to do things that are forbidden and hope they don't caught. Advertising companies also use street art in their ads because it gives people the impressions of youth and energy.  ${\bf \rule{1cm}{0.15mm}}_{4}$ Artists can show their pictures to an audience all over the world. Many city residents however say that seeing a picture on the Internet is never as good as seeing it alive. ${\bf \rule{1cm}{0.15mm}}_{5}$. There it will continue to change and grow 
         \\ \\ \textbf{Candidates:} \\ 
        A. Street art is a very popular form of art that is spreading quickly all over the world.  \\
        B. Today the Internet has a big influence on street art.  \\ 
        C. With the development of science and technology different art styles come into the Internet. \\
        D. The street art movement lives with the energy and life of a big city. \\
        E. People often have different opinions about street art. \\
        F. Street art used to be illegal but now has become popular. \\
        G. Some of them do not like artists who make so much money in galleries and museums.  \\
        \\
        \textbf{Answers with Reasoning Type}:\\ 1$\rightarrow$A (\textit{Summary}), 2$\rightarrow$E (\textit{Inference}), 3$\rightarrow$G (\textit{Inference}), 4$\rightarrow$B (\textit{Inference}), 5$\rightarrow$D (\textit{Inference}) (C and F are distractors)
        \\ \\
        \textbf{Discussion}: Blank 1 requires an answer which makes an overall broad statement to introduce the topic. Working backwards, this requires summarizing or finding a broad topic given the latter sentences.          
        \\
        \end{tabular}   \\ \specialrule{\cmidrulewidth}{0pt}{0pt} %\hline
        
    \end{tabular}
    \caption{Representative examples with diverse reasoning types}
    \label{tab:diverseReasoningExamples}
\end{table*}

\section{Additional Experiment Specifications}
\subsection{Specific BERT Model Used}
We use uncased BERT models for all our experiments. We use the BERT models trained by the canonical pytorch implementation of \newcite{Wolf2019HuggingFacesTS}.

\section{More examples}
We show more examples belonging to different reasoning categories in Table~\ref{tab:appendx_reason}. Also, some completed questions with strong distractors, multi-blank logic and diverse reasoning types are shown in Table~\ref{tab:strongDistractorExamples},~\ref{tab:multiBlankLogicExamples} and ~\ref{tab:diverseReasoningExamples}.

\end{document}